# Stream Computing

Subhash Kak

## Introduction

Stream computing is often seen to be embodied in compute-intensive kernel functions that are applied to each element in the data stream one at a time. These kernel functions operate in sequence in a pipelined fashion in what is essentially SIMD (single instruction multiple data) architecture. The advantage of doing this is the simplification of interconnects to get large increase in performance and in simplified programming. This processing is a take off on the style of computing that is found in DSP applications such as in voice, images, and video applied to a much larger range of applications.

Here we speak of stream computing in a much larger philosophical setting that is sometimes expressed in the idea of *stream* of consciousness. In literature, this idea is meant to imply the internal monologue that goes on in the mind. Such a stream has no single focus and it may shift from one to another, suggesting that the processing that lies behind it corresponds to a variety of processing centers that randomly project – in sequence – in the theater of the mind. But the motivation for this technical paradigm is the imperative of neuroscience more than the literary allusion.

The brain is composed of several modules each of which is essentially an autonomous neural network. Thus the visual network responds to visual stimulation and also during visual imagery, which is when one sees with the mind's eye. Likewise, the motor network produces movement and it is active during imagined movements. However, although the brain is modular, a part of it, located for most people in the left hemisphere, monitors the modules and interprets their individual actions in order to create a unified idea of the self. In other words, there is a higher integrative or interpretive module that synthesizes the actions of the lower modules [1].

Thus, if one were to construct a machine that is similar in organization to the brain, we would need a system of autonomic and parallel centers together with integrative processors that work at a higher level of abstraction to implement high level "intelligent" processing. It would be wise to use this framework to create new architectures to deal with high-volume data involving high-level reasoning.

As a caveat it must be said that this, in itself, will not endow the system with biological type of intelligence since another hallmark of biological intelligence that we are not in a position to simulate effectively in our implementations is that of reorganization with respect to changing environment [2-4].

From a practical implementation point of view, one can view stream computing as an overarching paradigm of which the current SIMD implementations of uniform stream processing are an elementary embodiment. The general streaming paradigm would have several SIMD machines operate in parallel on many copies of the data followed by further processing on another machine that operates at a higher abstraction. The new computing paradigm should make it possible to perform rapid processing of fast, high-volume data streams.

## The Path to Streaming

Classical computers are based on ideas that developed in the 1930s and 1940s to give shape to the intuition of how the rational mind performs computation. The general-purpose computing machine was visualized to consist of four main parts. These are the parts relating to the arithmetic logic unit, memory, control, and interface with the human operator.

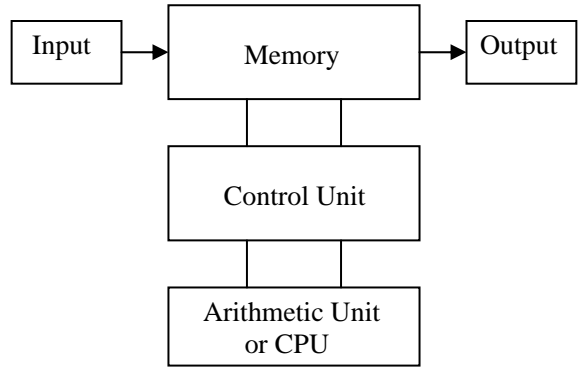

Figure 1: The stored-program computer architecture

The conception of the computer instrumented the formal notion of an algorithm. The innovative leap in building a general purpose device was the storing not only of data and the intermediate results of computation, but also the instructions that brought about the computation.

In the classical computer's memory there is no fundamental distinction between data and instruction, which is considered a shortcoming by some. Other claimed shortcoming are: the memory is monolithic and it must be sequentially addressed; it is single dimensional whereas in nature patterns of memory are multidimensional; and the attributes of data are not stored together with it, which is in contrast to what obtains in a higher level language where we expect a generic operation to take on a meaning determined by the meaning of its operands.

The reason why the architecture of the classical computer came to have this form is clear enough when we see it as an embodiment of serial computation carried out by the rational mind in arithmetic and other numerical tasks.

However, whereas some computations carried out by humans (especially those dealing with numerical computations) do fall within the category that is well captured by serial computation, there are a vast number of other computations that do not. In particular, tasks associated with "intelligence," which typically involves processing enormous amounts of data do not involve deliberate computation. In such tasks, autonomous centers appear to carry out computations independently, reducing the dimensions of the data and mapping it into an abstract space where further computations are done.

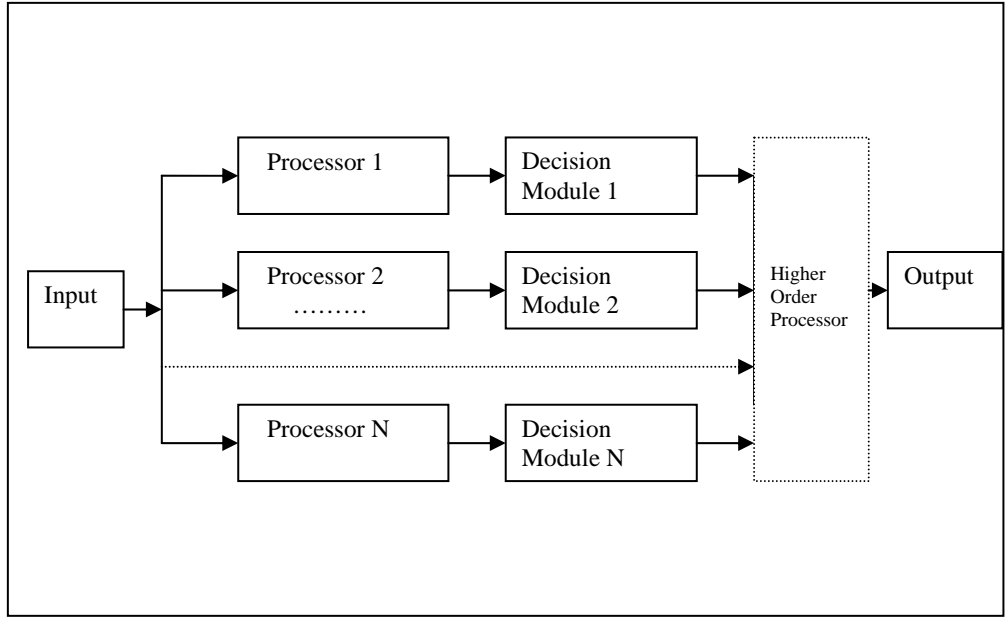

Figure 2: The stream computing framework

Although much of the computations are done in parallel, this is not the parceling out of computational tasks to different processors by taking advantage of the parallel components of the algorithm, which is what happens in what is technically called "parallel computing" [5]. Rather, here the entire data is seemingly *pushed* into a variety of autonomous processors, quite as a stream of water is pushed into various channels with different function, justifying the term *stream computing*. The higher-order processor cannot be generic and it must use specific application knowledge to design it.

**The Biological Context**

There is a wealth of experimental evidence from neuroscience that suggests that the conscious mind "creates" its reality in order to have a narrative that is "consistent" with the information reaching it from various specialized modules. This is seen most clearly in subjects who have suffered brain injury where the effect becomes most pronounced.

In the 60s and the 70s, Kornhuber and Deecke performed a series of experiments to measure the correlation between electrical activity in the brain (EEG) and a voluntary act. They found that the EEG from the area corresponding to the finger in the motor cortex for a subject who is about to move a finger starts to build up several hundred milliseconds before the conscious decision to make the act is made [6]. The conscious mind appears to label such an act its own free decision although one might dispute this.

Libet et al, in a variation of this experiment, showed that the EEG potential appeared to increase about 0.3 seconds before the subject made his "conscious choice" to flex his finger. These results are in agreement with the idea of the cortex constructs a model that is consistent with the mediating experience [7].

Likewise, blindsight subjects can see the completed Kanizsa triangle in their supposedly blind field. This indicates that the space of our visual experience is located outside of the visual cortex.

Subhash Kak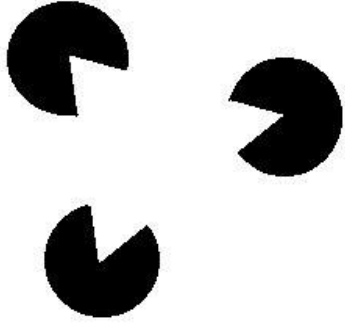

Figure 3: A Kanizsa triangle

Other examples of the conscious brain harmonizing the activity of various modules of the brain come from split-brain patients. In the words of Gazzaniga [8]:

> In a complication of stroke called anosognosia with hemiplegia, patients cannot recognize that their left arm is theirs because the stroke damaged the right parietal cortex, which manages our body's integrity, position, and movement. .. When patients with this disorder are asked about their arm and why they can't move it, they will say "It's not mine" or "I just don't feel like moving it"—reasonable conclusions, given the input that the left-hemisphere interpreter is receiving.
>
> The left-hemisphere interpreter is not only a master of belief creation, but it will stick to its belief system no matter what. Patients with "reduplicative paramnesia," because of damage to the brain, believe that there are copies of people or places. In short, they will remember another time and mix it with the present. As a result, they will create seemingly ridiculous, but masterful, stories to uphold what they know to be true due to the erroneous messages their damaged brain is sending their intact interpreter.

These findings do not fit into any simple model of computation, reminding us that there remains a big gulf between the reality of biological computing and our current explanations.

Other perspectives from which to look at information processing by the brain are those of learning and structure. Learning of memories is related to two distinct types: short-term and long-term. It is also believed that much of learning is based on recursive primitives. Recursion may be seen as the nesting of stories inside stories, dolls inside dolls, or the flowering of leaves and branches in a self-similar manner in a tree. The motor-sensory cortex maps the world as experienced by the senses and it is connected in a nested form to deeper layers.

Recursive representations are used implicitly or explicitly in scientific thought, literature, art and music. It is the basis of mental images not only of physical systems but also of behavior. In music, recursion helps in the creation of structure, as in the employing of a delayed version of a tune as its own accompaniment. In Indian music, rhythms and melodies are created out of set notes constituting the *raga*, which is then syncopated, translated, and rhythmically shrunk and enlarged to communicate emotion. Baroque composers have used recursion in their canons and fugues.

A single piece of music may begin as a melody. The rules by which the piece expands are that of recursive incorporation, replication, cleaving, and displacement, one transformation of the basic pattern nesting inside another. The notes of the basic melody are replicated several times throughout the piece, and incorporated within this replication may be its play at twice or half the speed and this may be assigned to different instrument voices. The core melody may be played at other times in retrograde and assigned to a different instrument to increase the beauty of the piece. The texture may change further with an inversion of everything that was done so far as in Johan Sebastian Bach's Musical Offering.

Other behavior, whether routine or creative, is also constructed of similar primitives.

**Learning**

Learning requires networks that have the capacity to generalize very quickly since the decision time is short. It was the motivation to model short-term memory that led to the development of the corner classification (CC) family of the instantaneously trained neural networks (ITNN) [9-11] that learn instantaneously and have very good generalization performance. ITNN networks are an example of prescriptive learning; they train the network by isolating the corner in the n-dimensional cube of the inputs represented by the input vector being learnt. In specific modules, the processors may have strong feedback connections [12-15].

Another approach to learning is that of using recursive structures which are implicitly mapped to the recursive primitives of the information process. This approach which is called the Network of Networks (NoN) approach is based on a model of cortical processing. It is based on two fundamental ideas in neurobiology: (1) groups of interacting nerve cells, or neurons, encode functional information, and (2) processing occurs simultaneously across different levels of neural organization. Although the NoN model makes simplifying assumptions about neural activity, systems using this approach have performed very well. Nested distributed systems denote the core architecture of the model. In the context of describing computational features of the cerebral cortex, this organizing principle was proposed by Mountcastle [16] and applied by Sutton and Anderson [17].

The formation of clusters and levels among neurons is based on their interconnections. The NoN model suggests that despite enormous diversity in the connection patterns associated with individual neurons, many neural circuits can be subdivided into essentially similar sub-circuits, where each sub circuit contains many types of neurons. This hierarchy is evident in the cerebral cortex, which has long been considered the most complex and elusive of neural circuits. (See Figure 4 adapted from [17]). This nesting arrangement serves to link different and often widely separated regions of the cortex in a precise but distributed manner. Several physiological responses, such as those occurring in the visual cortex in response to optical stimuli, may be associated with each first-level sub-circuit.

Regarding the question of universals underlying aesthetics in art and music [18], it has been argued that although there are no specific universals that have a quantitative form, there are qualitative ones that are ultimately related to the neurophysiological basis of the cognitive system. These qualitative universals also have a recursive nature.

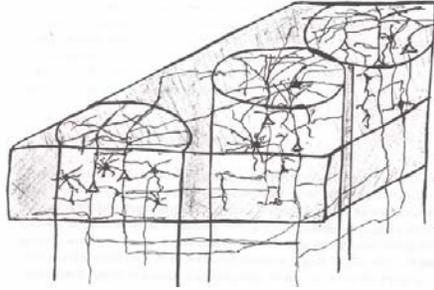

Figure 4: Schematic representation of the cerebral cortex. Three networks of intermediate level organization are displayed, showing recursive structure.

**Applications**

Clearly stream computing is a paradigm whose implementations embody elements that are appropriate to the specific application. In practice, it would consist of several layers of SIMD processors working to abstract different application-specific abstractions from several copies of the data. This should make it possible to do faster processing and analysis as well as decision making in business and science. It should also allow one to deal with data in a variety of forms that include transaction and internet data, video, voice, and data from sensors.

The processing of data will begin as it streams in and, therefore, it must use models and abstractions that can deal with incomplete information. This is to be contrasted from conventional approaches where the data is first collected and stored and then analyzed. Clearly, the conventional approach takes time that might not be available in applications such as automated trading or in surveillance.

One can visualize applications of stream computing include those to life sciences (as in computational biology, computational immunology and visualization), automated trading (as is done by hedge funds), and various types of graphics-based consumer applications. Stream computing should also be of importance in security applications, data mining, and intrusion detection.

Department of Computer Science, Oklahoma State University, Stillwater, OK 74078, U.S.A.